\pdfoutput=1
\documentclass[11pt]{article}

\usepackage[]{ACL2023}

\usepackage{times}
\usepackage{latexsym}
\usepackage[most]{tcolorbox}

\usepackage[T1]{fontenc}
\usepackage[utf8]{inputenc}

\usepackage{microtype}

\usepackage{inconsolata}

\usepackage{graphicx}
\usepackage{booktabs}
\usepackage{xcolor}
\usepackage{colortbl}
\usepackage{multirow}
\usepackage{balance}

\definecolor{bondiblue}{rgb}{0.0, 0.58, 0.71}
\definecolor{headercolor}{rgb}{0.9, 0.9, 0.9}

\definecolor{bondiblue}{rgb}{0.0, 0.58, 0.71}

\usepackage{graphicx}
\usepackage{subcaption}
\usepackage{xcolor}
\usepackage{balance}

\newcommand{\gtlogo}{\raisebox{3.4pt}{\includegraphics[scale=0.017]{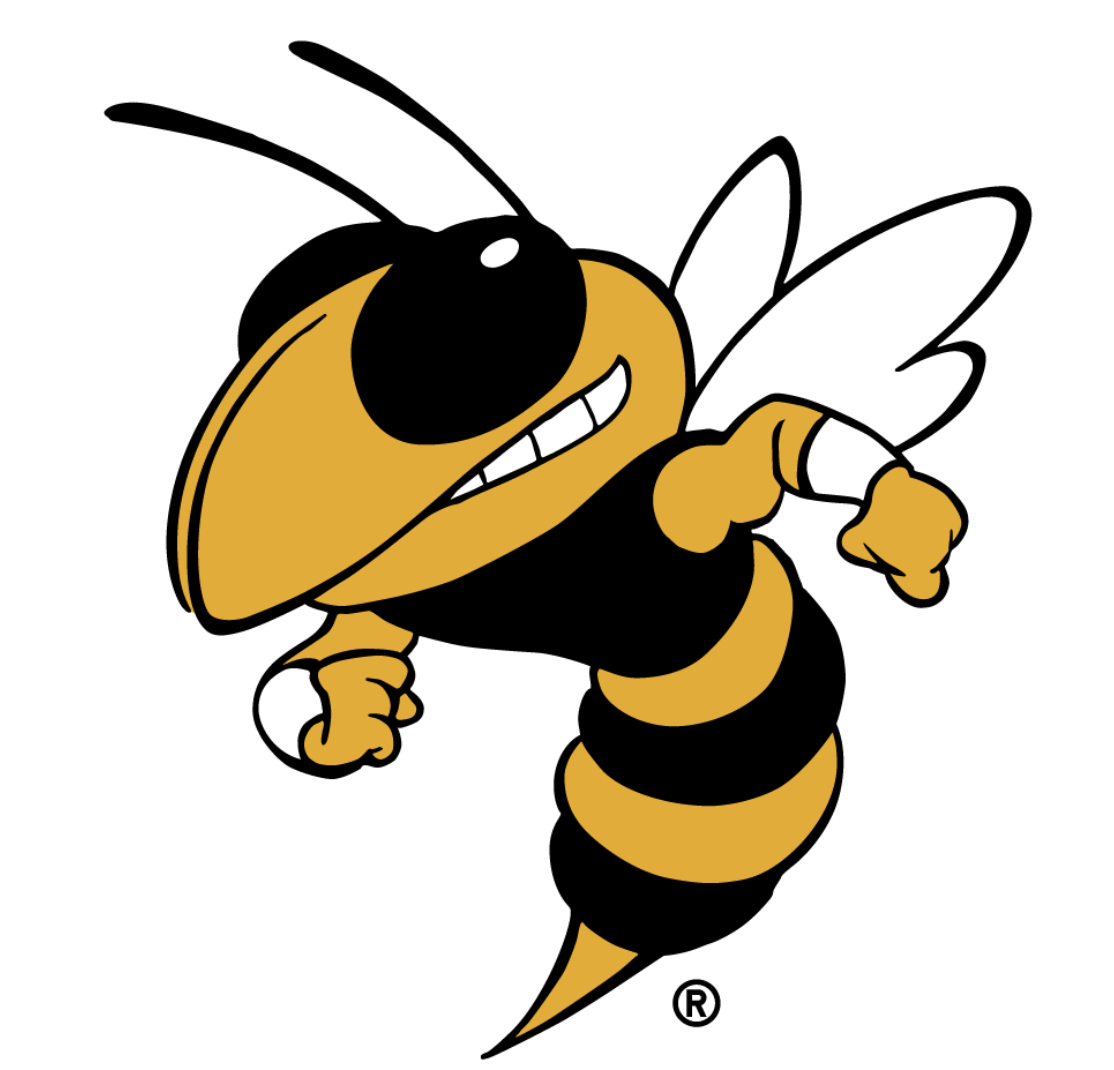}}}
\newcommand{\adllogo}{\raisebox{3.4pt}{\includegraphics[scale=0.012]{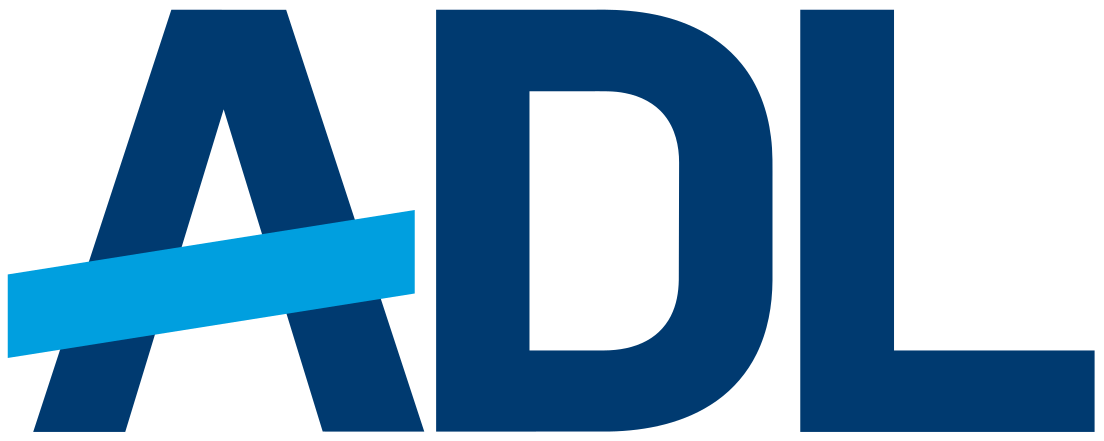}}}
\newcommand{\gt}{\gtlogo}
\newcommand{\adl}{\adllogo}

\title{A Community-Centric Perspective for Characterizing and Detecting\\ Anti-Asian Violence-Provoking Speech}

\author{Gaurav Verma \gt \hspace{1.5em}
        Rynaa Grover \gt \hspace{1.5em}
        Jiawei Zhou \gt \hspace{1.5em}
        \textbf{Binny Mathew} \adl \\
        \textbf{Jordan Kraemer} \adl \hspace{1.5em}
        \textbf{Munmun De Choudhury} \gt \hspace{1.5em}
        \textbf{Srijan Kumar} \gt \hspace{1.5em} \\
        \gt Georgia Institute of Technology, \adl Anti-Defamation League\\
        \texttt{\{gverma, rgrover30, j.zhou\}@gatech.edu}\\
        \texttt{binny.iitkgp@gmail.com}, \texttt{jkraemer@adl.org}\\
        \texttt{munmun.choudhury@cc.gatech.edu}, \texttt{srijan@gatech.edu}
        }

\begin{document}
\maketitle
\begin{abstract}
Violence-provoking speech – speech that implicitly or explicitly promotes violence against the members of the targeted community, contributed to a massive surge in anti-Asian crimes during the COVID-19 pandemic. While previous works have characterized and built tools for detecting other forms of harmful speech, like fear speech and hate speech, our work takes a community-centric approach to studying anti-Asian violence-provoking speech. Using data from $\sim420k$ Twitter posts spanning a 3-year duration (January 1, 2020 to February 1, 2023), we develop a codebook to characterize anti-Asian violence-provoking speech and collect a community-crowdsourced dataset to facilitate its large-scale detection using state-of-the-art classifiers. We contrast the capabilities of natural language processing classifiers, ranging from BERT-based to LLM-based classifiers, in detecting violence-provoking speech with their capabilities to detect anti-Asian hateful speech. In contrast to prior work that has demonstrated the effectiveness of such classifiers in detecting hateful speech ($F_1 = 0.89$), our work shows that accurate and reliable detection of violence-provoking speech is a challenging task ($F_1 = 0.69$). We discuss the implications of our findings, particularly the need for proactive interventions to support Asian communities during public health crises.\\
\noindent\textcolor{red}{\textit{Warning: this paper contains content that may be offensive or upsetting.}}
\end{abstract}

\section{Introduction}
\label{sec:introduction}
Online platforms struggle with various forms of information pollution, including but not limited to harmful speech and misinformation. These malicious phenomena often intertwine in complex ways, exacerbating real-world issues. A glaring example of this is the dramatic increase in hate crimes against Asian communities during the COVID-19 pandemic, which included physical assaults and verbal harassment~\cite{nyt2021asian-americans}. Rumors about the virus's origins coalesced with pre-existing prejudices, resulting in narratives that portrayed Asians as ``uncivilized'', blamed them for the virus, and labeled them as ``spies''. This intricate interplay between different forms of malicious content led to a 339\% surge in anti-Asian crimes in 2021~\cite{nbcnews2023anti-asian}. Such narratives have had a far-reaching impact, making community members afraid to engage in basic daily activities, from grocery shopping to using public transit~\cite{nyt2021selling}. Moreover, this increase in real-world attacks was not limited to one ethnic group but affected diverse Asian communities, including Chinese, Korean, Vietnamese, and Filipino Americans~\cite{nyt2021asian-americans}.

In light of the dramatic uptick in anti-Asian violence and ensuing community fear, the necessity to accurately identify \textit{violence-provoking speech} --- i.e., \textit{speech that could promote real-world violence against members of targeted communities} ~\cite{benesch2021dangerous}, becomes paramount. This differs from \textit{hateful speech}, which is \textit{a more subjective form of expression that may not directly incite violence}~\cite{benesch2021dangerous}. While both these phenomena share commonalities --- being rooted in prejudice and derogation --- violence-provoking speech constitutes a specific \textit{subset} of hateful speech that explicitly or implicitly encourages acts of aggression. The higher severity of harm associated with violence-provoking speech~\cite{scheuerman2021framework} calls for targeted approaches for its detection, beyond treating hate as a monolithic entity. 

Recognizing that the perception of \textit{what} qualifies as violence-provoking is not universal but varies among the targeted communities, we adopt a \textit{community-centric} approach. We leverage the ``insider perspectives''~\cite{kim2021you} and the subjective lived experiences~\cite{dredge2014cyberbullying} of community members to capture the nuances, slurs, and coded language that may be overlooked by outsiders. Our focus is particularly on Asian communities in the context of the COVID-19 pandemic, who were disproportionately impacted by violence-provoking speech leading to real-world harm.
We address two key research questions:

\vspace{0.01in}
\noindent\textbf{RQ1}: \textit{What are the characteristics of violence-provoking speech that targets Asian communities? How is anti-Asian violence-provoking speech different from anti-Asian hateful speech?}\\
\noindent\textbf{RQ2}: \textit{Can state-of-the-art natural language processing (NLP) approaches accurately detect violence-provoking speech? How do the detection abilities compare to that of hate speech detection?}

\vspace{0.01in}
{We address these research questions by developing and validating a codebook for identifying anti-Asian violence-provoking speech, while working with Anti-Defamation League, a leading non-governmental organization that specializes in tackling real-world hate and extremism (\textit{RQ1}).} We then use the codebook to obtain crowd-sourced annotations for violence-provoking and hateful content from individuals who self-identify as Asian community members. Our dataset demonstrates high inter-rater agreement (Fleiss' $\kappa$ = 0.66 for violence-provoking speech labels).\footnote{Project webpage: \url{https://claws-lab.github.io/violence-provoking-speech/}}

We then use the annotated data to develop binary classifiers that are trained to distinguish \textit{(i)} violence-provoking content from not-violence-provoking content and \textit{(ii)} hateful content from non-hateful content. We find that while the NLP approaches effectively detect hateful speech ($F_1$ score = $0.89$), it is relatively more challenging to detect violence-provoking speech ($F_1$ score = $0.69$) (\textit{RQ2}), perhaps due to its nuanced and subjective nature that relies on victims' own perceptions.

We discuss possible reasons why detecting violence-provoking speech is challenging and the implications of lacking capabilities of the detectors. Additionally, we discuss how our developed approaches could aid in the development of moderation algorithms that employ tiered penalties for content that violate norms of varying severities. Finally, we highlight the need to develop trauma-informed approaches to proactively support targeted communities.

\section{Related Work}
\label{sec:related_work}
We categorize relevant prior work into three categories: (i) studies focusing on different forms of harmful speech on online platforms, including fear speech and hateful speech, (ii) studies of anti-Asian content, and (iii) detection methods for different forms of harmful speech.

\vspace{0.01in}
\noindent\textbf{Forms of harmful speech}:
Prior studies on harmful speech have mostly focused on instances of hate speech. ~\citeauthor{ezeibe2021hate} (\citeyear{ezeibe2021hate}) show how hate speech is an often neglected driver for election violence. Another study by \cite{williams2020hate} has shown that online hate victimization is part of a wider process of harm that can begin on social media and then migrate to the physical world.

Another form of harmful speech is fear speech. According to ~\citeauthor{buyse2014words} (\citeyear{buyse2014words}), Fear speech is defined as an ``expression aimed at instilling (existential) fear of a target (ethnic or religious) group.'' While it cannot be pinpointed if fear speech can cause violence, it can lower the threshold for violence~\cite{saha2021short}. ~\citeauthor{saha2023rise} (\citeyear{saha2023rise}) study the prevalence of fear speech in a loosely moderated community (Gab.com) and observe that users posting fear speech are more influential as compared to those who post hate speech. The authors argue that this is mainly due to the nontoxic and argumentative nature of the speech posts. Violence-provoking speech is closely related to fear speech but is defined more specifically as speech that promotes violence. Closest to the concept of violence-provoking speech, ~\citeauthor{benesch2012dangerous} (\citeyear{benesch2012dangerous}), define dangerous speech as an expression that has ``a significant probability of catalyzing or amplifying violence by one group against another, given the circumstances in which they were made or disseminated.'' 
In this work, we contextualize the framework by ~\citeauthor{benesch2012dangerous} (\citeyear{benesch2012dangerous}) for anti-Asian violence-provoking speech by using a community-centric approach and aim to operationalize its large-scale detection using state-of-the-art classifiers.

\vspace{0.01in}
\noindent\textbf{Anti-Asian hate}:
The outbreak of the COVID-19 pandemic has led to the spread of potentially harmful rhetoric, conspiracy theories, and hate speech towards several Asian communities. ~\citeauthor{tahmasbi2021go} (\citeyear{tahmasbi2021go})  collected two large datasets from Twitter and Reddit(/pol/) and observed that COVID-19 was driving the rise of Sinophobic content on the web. While counterspeech users were actively engaged with hateful users, users were highly likely to become hateful after being exposed to the hateful rhetoric~\cite{he2021racism}. In this work, we specifically focus on anti-Asian content that has the potential to provoke violence.

\vspace{0.01in}
\noindent\textbf{Detection methods}: \citet{elsherief-etal-2021-latent} propose a taxonomy of implicit hate speech and consider factors, including incitement to violence and intimidation. They also investigate the use of BERT-based classifiers for detecting implicit hate speech and discuss the underlying challenges like models struggling with coded hate symbols and entity framing. More recently, with advent of large language models (LLMs) like GPT-4, \citet{matter2024close} found good agreement between GPT-4 annotation's and human coders in identifying violent speech. In this work, we aim to study if the capabilities of NLP classifiers also translate to anti-Asian violence-provoking speech that is curated in a community-crowdsourced manner. 

\section{Study Overview}
\label{sec:study_overview}

\begin{figure*}[!t]
    \includegraphics[width=1.0\textwidth]{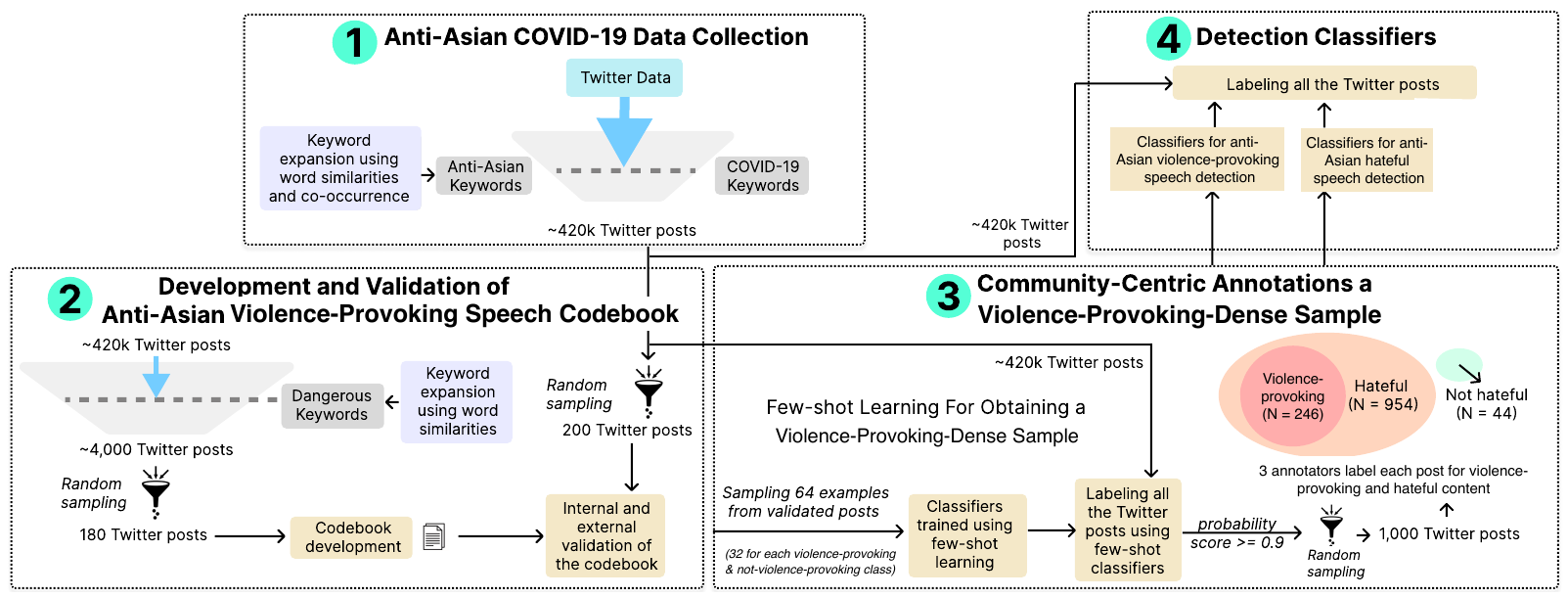}
    \caption{{\textbf{Study Overview.} Our study comprises 4 key parts: \textit{(i)} collecting anti-Asian COVID-19 data from Twitter, \textit{(ii)} developing and validating anti-Asian violence-provoking speech codebook, \textit{(iii)} obtaining community-centric annotations, and \textit{(iv)} training and evaluating detection classifiers on community-crowdsourced data.}}
    \label{fig:data-pipeline}
\end{figure*}

Our study involves collecting a large-scale Twitter dataset that comprises anti-Asian content related to the COVID-19 pandemic. It is worth noting that given the harmful impact of violence-provoking speech, it is challenging to find examples that explicitly provoke violence, as they are frequently removed from the platform due to moderation efforts. However, instances of such speech still exist on major platforms. To overcome the challenge of collecting potentially violence-provoking speech examples from Twitter, we designed an elaborate pipeline to obtain a dense sample, which we then used for obtaining annotations from the Asian community members. 

As shown in Figure \ref{fig:data-pipeline}, there are 4 key parts to our study. The first part comprises data collection, where we collected about $420k$ Twitter posts containing anti-Asian keywords and COVID-19-related keywords. We then used posts in subsequent parts of our study to answer our RQs:\\
\noindent\textbullet\hspace{2pt} To facilitate the effective development of the anti-Asian violence-provoking speech codebook (\textit{RQ1}), we prepared a set containing a reasonably substantial amount of potentially harmful content. The original $\sim420k$ Twitter posts comprise a variety of topics, such as hateful expressions, counter to hateful expressions, or instances of someone sharing anecdotes about anti-Asian hate. However, since we are particularly interested in violence-provoking speech --- speech that implicitly or explicitly promotes violence against the Asian community, we further filtered down the collected $\sim420k$ Twitter posts to find a set of $4,076$ Twitter posts that is denser in violence-provoking content. This concentrated sample is essential for the manual analysis required in codebook development, without the impractical inundation of non-violence-provoking posts that risk overburdening annotators in a task that involves reviewing potentially traumatizing content. In Section \ref{sec:codebook},  we describe how we used this subset of Twitter posts to develop a codebook for anti-Asian violence-provoking speech while emphasizing its distinction from hateful speech.\\
\noindent\textbullet\hspace{2pt} Our next goal was to collect annotations from community members to train, evaluate, and contrast classifiers for violence-provoking and hateful speech (\textit{RQ2}). To ensure the wide applicability of the classifiers, the annotated Twitter posts should cover diverse Twitter posts, which may not be included in the $4,076$ posts filtered using the identified `dangerous' keywords. At the same time, platform moderation makes a random sample from the $\sim 420k$ sparse in violence-provoking content. To this end, to curate a sample of Twitter posts with high diversity and a higher density of violence-provoking posts, we used a prompt-based few-shot learning approach. Few-shot learning in natural language processing has not only demonstrated great effectiveness in terms of data-efficient accuracy~\cite{schick2021exploiting, mozes2023towards} but also in terms of generalizability over out-of-domain distributions~\cite{liu2022sample}. Using the few-shot classifier, we selected a subset of $1,000$ of Twitter posts that are potentially dense in violence-provoking speech content. We then obtained community-centric speech annotations for these posts (Section \ref{sec:few-shot-learning}), and used them to train and evaluate classifiers for violence-provoking and hateful speech detection (Section \ref{sec:hate_and_danger_classifiers}). We contrasted the classifiers' capabilities and conducted error analysis to understand their shortcomings.

\section{Anti-Asian COVID-19 Data Collection}
\label{sec-antiasian-data}
We started by collecting a large-scale dataset from Twitter using COVID-19 and anti-Asian keywords. We used the Twitter Academic API to collect data from January 1, 2020 to February 1, 2023, by querying using COVID-19 and anti-Asian keywords together. To consider a wide range of data, we expanded the set of keywords used in prior work.

\vspace{0.05in}
\noindent\textbf{Keyword expansion strategies}: The present study commenced with an initial set of $6$ COVID-19 and $16$ Anti-Asian keywords, adapted from \citet{he2021racism}. We removed certain keywords like `ccpvirus' as the focus of our work is on speech that targets Asians and not speech that may involve political factors.  Based on the initial keyword set, we obtained $16 \times 6 = 96$ combinations of keywords. For each unique combination of anti-Asian and COVID-19 keywords, we queried Twitter to collect posts that contain both keywords. Since anti-Asian speech could have evolved since \citeauthor{he2021racism} (\citeyear{he2021racism}) conducted their study, we adopted two strategies to further expand the list of anti-Asian keywords.

\vspace{0.02in}
\noindent\textbf{(i) Word co-occurrence}: We calculated the similarity scores between pairs of words based on their co-occurrence frequency. The intuition behind this approach is that anti-Asian words would co-occur frequently in the same post. For each initial keyword, two authors manually verified the top 5 co-occurring words and expanded the set.

\vspace{0.02in}
\noindent\textbf{(ii) word2vec similarity}:  We trained a word2vec~\cite{mikolov2013efficient} model on the dataset collected using the initial set of keywords, setting the embedding dimension to 100 and the context window size to 5. We then computed the cosine similarity between the words in the vocabulary and the initial list of anti-Asian keywords. Two authors then manually verified the top five similar words for each initial anti-Asian keyword and expanded the set of keywords.

We conducted this expansion process in a snowball fashion, repeating each approach five times to arrive at the final list of keywords. In each run, new keywords were identified using the keywords from the previous iteration, which were then manually verified to be relevant. We show the final list of $33$ anti-Asian and COVID-19 keywords used for our study in Table \ref{tab:final_keywords} (Appendix). We did another round of data collection by taking unique combinations of COVID-19 and expanded anti-Asian keywords as queries. Finally, we curated a dataset comprising $418,999$ Twitter posts.

\section{Development of Anti-Asian Violence-Provoking Speech Codebook}

\label{sec:codebook}
The existing guidelines developed by the Dangerous Speech Project~\cite{benesch2021dangerous} provide a general framework for identifying violence-provoking speech. We followed this guideline and expanded it to develop a comprehensive codebook that allows empirical measurement and categorization. We grounded the defined sub-concepts in real data from Twitter and contextualized them in the community-targeting  (i.e., anti-Asian) framework. To enable effective development of the codebook by qualitatively coding the data, we started by obtaining a subset of the $\sim 420k$ tweets by filtering based on `dangerous' keywords so that it is potentially concentrated in violence-provoking speech. 

\vspace{0.01in}
\noindent \textbf{Preparing a concentrated sample for codebook development}: 
To obtain dangerous keywords, we started with the example phrases mentioned in the practical guide provided by the Dangerous Speech Project~\cite{benesch2021dangerous} and expanded the phrases by computing similarity scores with the phrases in our dataset. For similarity computation, we used word2vec embeddings fine-tuned on our corpus. Table \ref{tab:dangerous_keywords} (Appendix) shows the initial dangerous keywords obtained from the Dangerous Speech Project and the expanded set after manually removing irrelevant and redundant phrases. It is worth noting that dangerous keywords include explicitly violence-provoking terms like `kill' as well as implicit keywords that indicate dehumanization (like comparisons to `ants,' `lice,') and the use of words like `mercy,' `charity,' and `forgive' to display virtuousness over the targeted community.  Of the $\sim420k$ Twitter posts, $4,076$ Twitter posts contained one or more dangerous phrases. We use this potentially rich sample in anti-Asian violence-provoking content for codebook development.

\vspace{0.01in}
\noindent \textbf{Developing violence-provoking speech codebook}: Our theory-guided, data-centric approach was inspired by previous work in understanding minority stress ~\cite{saha2019language} and credibility indicators ~\cite{zhou2023synthetic}. To operationalize the framework proposed by ~\citet{benesch2021dangerous} within the anti-Asian hate context and for empirical tasks, three of the authors first engaged in a discussion concerning its applicability and the potential need for expansions or alterations.
The discussion was informed by the authors' lived experiences, with some being members of the Asian community. 
Then, through an iterative process, the authors inductively developed categorical codes to characterize violence-provoking speech and deductively assessed the applicability. Two authors independently coded $180$ randomly sampled instances that contained dangerous phrases and drafted concepts with definitions and examples. With a third author involved, the discrepancies in annotations were discussed, and drafted categories were combined and revised. We tested this revised codebook with the sample until no new themes emerged, resulting in a unified and coherent document that reflected the perspectives and consensus of all contributors. 
{The codebook was then reviewed by all co-authors before being validated by members of the Asian community and experts in Anti-Defamation League (details in the following subsection).} Table \ref{tab:codebook} in Appendix shows the final codebook to characterize anti-Asian violence-provoking speech.

\begin{table}[!t]
    \centering
    \definecolor{headercolor}{rgb}{0.9, 0.9, 0.9}
    \resizebox{0.48\textwidth}{!}{
    \begin{tabular}{p{0.20\textwidth}  p{0.23\textwidth}  p{0.11\textwidth}}
        \toprule
        \rowcolor{headercolor} \textbf{Annotators} & \textbf{Violence-provoking} & \textbf{Hateful}\\ \toprule
        Internal Annotators & \hspace{8pt} $\kappa = 0.78$ & $\kappa = 0.86$\\
        External Annotators & \hspace{8pt} $\kappa = 0.69$ & $\kappa = 0.82$\\ 
        Prolific Annotators & \hspace{8pt} $\kappa = 0.66$ & $\kappa = 0.72$ \\ \bottomrule
    \end{tabular}}
    \caption{{\textbf{Codebook validation.} Fleiss' $\kappa$ scores for violence-provoking and hateful labels.}}
    \label{tab:annotations}
\end{table}

\noindent\textbf{Validating the codebook}: We validated the effectiveness of the codebook through pilot testing with both internal and external annotators. We randomly sampled $100$ tweets from the entire Twitter dataset, and for each tweet, the annotators were asked if it contained violence-provoking speech or hateful content.
We quantified the level of agreement between two annotators using Fleiss' $\kappa$. Internal testing was done by the two researchers who drafted initial codebooks, which demonstrated substantial agreement for both categories (Table \ref{tab:annotations}).
Following this, to establish the external validity of the codebook, we recruited two graduate students who identified as members of the Asian community and had no other involvement in the presented research. They annotated the same $100$ tweets, utilizing the codebook as a framework to determine the presence of violence-provoking and hateful content targeting the Asian community. Moreover, they were requested to provide a rationale outlining their assessment of the codebook's efficacy. The findings of the external validity evaluation revealed that the codebook exhibited strong external utility, underscored by the high Fleiss' $\kappa$ scores for both variables of interest (Table \ref{tab:annotations}). {We also incorporated suggestions from Anti-Defamation League\footnote{\url{https://www.adl.org/}}, a leading non-governmental organization that specializes in countering hate and extremism, to ensure that the codebook covers various aspects of anti-Asian violence-provoking speech. We used the feedback provided by the graduate students and the NGO partners to make minor changes to the codebook, primarily to rephrase certain definitions.}

\section{Community-Centric Annotations with a Violence-Provoking-Dense Sample}
\label{sec:few-shot-learning}
With the validated codebook, we aimed to get community-centric annotations for $1,000$ Twitter posts that contained diverse examples and were also rich in violence-provoking content. To enable this, we trained a few-shot classification model to filter relevant posts, given the remarkable generalizability and accuracy of few-shot learning. 

\vspace{0.01in}
\noindent\textbf{Prompt-based few-shot learning for creating a dense violence-provoking sample for community-centric annotation}: From the Twitter posts that were labeled while validating the codebook, we used a representative subset of labeled examples (a total of $64$ examples, $32$ were violence-provoking, and $32$ were not) to train a binary classification model. More specifically, we used a prompt-based few-shot learning method called Pattern Exploiting Training (PET)~\cite{schick2021exploiting}. This method converts the input data into cloze-style statements and fine-tunes a pre-trained language model to predict the missing token. We formulated the classification task using the following prompt: \texttt{Is "<input sentence>" violence-provoking? [MASK].}, with the \verb|[MASK]| token serving as a placeholder that could be verbalized as ``Yes'' or ``No'' during model training. We used the pre-trained \texttt{DeBERTa-v2-xxlarge} language model~\cite{he2020deberta} as the backbone and fine-tuned it for $200$ iterations using a batch size of $16$. The rest of the hyper-parameters were set to the values used in prior work~\cite{schick2021exploiting}. All the language models in this work were trained on an NVIDIA Quadro RTX 8000; the cumulative training time amounted to about $24$ GPU hours.   We then used the trained model to predict labels and prediction scores over the entire $\sim420k$ Twitter posts in our curated dataset. We then randomly selected $1000$ Twitter posts that had a classification score of $\geq 0.9$. Since these Twitter posts have a high probability score of being violence-provoking, the set of $1000$ Twitter posts is likely to contain a higher density of violence-provoking posts than the entire set of $\sim420k$ posts (labeling of $200$ examples from the filtered set by the authors indicated that about $30\%$ posts were violence-provoking). Furthermore, the generalizability of the few-shot learning approach helped in identifying violence-provoking posts that do not necessarily contain dangerous keywords. For instance, ``these are not people of god, they are virus-ridden ch**ks who do not deserve our goodness" does not contain any dangerous keywords in our list (see Table \ref{tab:dangerous_keywords}) but was correctly identified to be violence-provoking by the few-shot learning approach. Next, we used these $1000$ Twitter posts to obtain community-centric annotations.

\vspace{0.01in}
\noindent\textbf{Obtaining community-centric annotations}: We designed a custom interface to collect data from a crowd-sourcing platform called Prolific (\url{https://www.prolific.com/}). We recruited participants who identified as Asians, were located in the United States or Canada, spoke fluent English, and were at least 18 years old. The participants were compensated at an hourly rate of 10 USD per hour. To ensure high-quality annotations, we only considered participants who had at least $100$ submissions and had a minimum of $95\%$ approval rate across their past submissions. The overall cost of obtaining annotations was about $760$ USD. 

\begin{figure}[!t]
    \centering
    \includegraphics[width=1.0\linewidth]{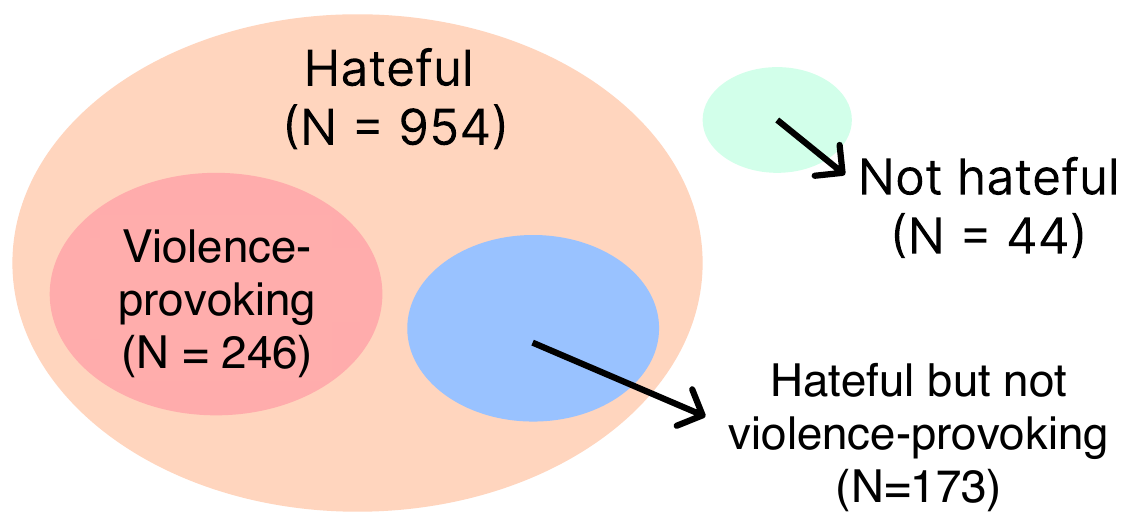}
    \caption{{Category-wise distribution of the dataset obtained from community-centric annotations. `Data statistics' in Section \ref{sec:few-shot-learning} provides additional details.}}
    \label{fig:enter-label}
\end{figure}

\begin{table*}[!h]
\centering
% \small 
\resizebox{1.0\textwidth}{!}{%
\definecolor{headercolor}{rgb}{0.9, 0.9, 0.9}
\begin{tabular}{p{20cm}} %{p{1.52\textwidth}}
\toprule
\rowcolor{headercolor} \textbf{violence-provoking and hateful (N = 246)} \\
\textbullet\hspace{2pt} f**ken slanteyed p*gs brought the virus in sa yet y’all post black hands. f**king racist weak p*gs \\
\textbullet\hspace{2pt} they had it all along trust me, can’t trust the asian spies here. they helped spread the virus \\
\textbullet\hspace{2pt} f**king d*g eat**s, asian wea**s don’t have balls to fight like a man that’s  why the pandemic \\
\textbullet\hspace{2pt} clearly your a hater a democrate or maybe a communist like the communist chinese who launched lied threw a deadly virus at the world these ch**ks must pay for this evil jealous liars and there will be 4 moreee years trumpster.. 4 moreeee years trumpster \\
\textbullet\hspace{2pt} they don’t care about schools or kids! these are sat**ic baby eatin soulless traitors who are at war with america on our soil. bet if you lined em ch**ks up and s**t em all this corona would be over yesterday\\
\midrule
\rowcolor{headercolor} \textbf{Hateful but not violence-provoking (N = 173)} \\
\textbullet\hspace{2pt} open them eyes up ugly c**nk b**ch. you wish you looked as good as me \#covid\\
\textbullet\hspace{2pt} they are giving them the virus and sending to africa. them ch**ks playing chess on us \\
\textbullet\hspace{2pt} manmade chinese virus is designed to kill anyone in the world who is not a c**nk \\
\textbullet\hspace{2pt} he didn’t k**l them chi**s, a virus did. c**nks. \\
\textbullet\hspace{2pt} she don’t have ch***y eyes, girl got big **s eyes just like her paw. she aint getting the virus\\
\midrule
\rowcolor{headercolor} \textbf{Not violence-provoking and not hateful (N = 44)} \\
\textbullet\hspace{2pt} this is why we hate wasians like him. white passing bi***es like him don't get called c**nk . \\
\textbullet\hspace{2pt} @user you people think he can make miracles. blame the chinese for the wuhan kung flu virus. what the hell would you do. liberals only know how to create riots looting and thugs \\
\textbullet\hspace{2pt} people do bad and when they get the same back they play victims. chinese 3 weeks ago wherever they existed were segregated, seen as virus carriers, treated as if they were sub humans things have since changed and the chinese are doing the same, and now we are all screaming  with squinted eyes..\\
\textbullet\hspace{2pt} when stupidity prevails over intelligence. now the deads and the spread of the chinese virus killing more people gona be responsibility of that federal stupid and slanted judge\\
\textbullet\hspace{2pt} it is a weapon that's why china lied about it they don't care about their people they killed 70 million to get in power so they released it on they're none party members \#covid \#gooks\\
\bottomrule
\end{tabular}%
}
\caption{\textbf{Qualitative examples belonging to different subsets of the data annotated by the community members.}}
\label{tab:qualitative_examples}
\end{table*}

Our user-friendly interface required the eligible participants to read the consent form and acknowledge their consent and eligibility before being directed to the codebook that contained all the necessary definitions and examples (see Table \ref{tab:codebook} in Appendix). We instructed the participants to read the codebook thoroughly. Once the participants had confirmed reading the codebook, they were directed to an `Examples page' where they were shown $5$ different illustrative Twitter posts and were asked to provide annotations for the following questions: ``\textit{Does this post contain violence-provoking speech that targets Asian community member(s)?}" and ``\textit{Does this post contain hateful speech that targets Asian community member(s)?}"; for each of the questions the participants could respond in a `yes'/`no'.  Upon responding to the examples, the participants were shown the expected answers and the rationale for the expected categorization. The participants could revise their response to the example posts multiple times, and only after they had correctly answered the 5 examples were they directed to the annotations page. On the annotations page, the participants saw $25$ posts, one by one, for which they provided answers to the questions above in yes/no. The annotations page also contained an abridged version of the codebook.

The participants could discontinue the study at any point for any reason while still getting compensated for the time spent on the study. We obtained annotations from $3$ different participants for each of the $1,000$ Twitter posts. In all, we received annotations from $120$ unique participants, with a median age of $31$, among which $54\%$ identified as males and the rest as females. The country of birth for the participants was distributed across the United States, Canada, China, Korea, Philippines, Japan, and Vietnam. On average, a participant took 29 minutes to complete one session.

\vspace{0.01in}
\noindent\textbf{Data statistics:} We assigned final labels for violence-provoking and hateful speech based on majority annotations. In other words, if at least 2 out of 3 annotators considered a Twitter post to be either violence-provoking (or hateful), it was assigned as violence-provoking (or hateful). Out of $1000$ Twitter posts, $246$ tweets were classified as both hateful and violence-provoking, and $954$ tweets were hateful but not necessarily violence-provoking. We only considered a Twitter post to be \textit{not violence-provoking} when it received \textit{no} annotations (i.e., 0 out of 3) for the violence-provoking label to avoid any ambiguity.  Adopting this definition, $173$ Twitter posts were identified to be \textit{hateful but not violence-provoking}. Similarly, if a Twitter post received $0$ out of $3$ annotations for being hateful, it was considered not hateful ($n = 44$). We present some qualitative examples from the community-annotated data in Table \ref{tab:qualitative_examples} and a visualization of the categorical distribution of data in Figure \ref{fig:enter-label}.  Finally, we observed that the inter-annotator agreement between the annotators, as quantified by Fleiss' $\kappa$, was $0.66$ and $0.72$ for violence-provoking and hateful labels, respectively (see Table \ref{tab:annotations}), which are notably better than the agreement scores reported in prior related studies~\cite{saha2021short}. 
Next, we use the curated dataset to develop classifiers that can detect violence-provoking and hateful speech by fine-tuning pre-trained language models for respective binary classification tasks. 

\section{Contrasting Hateful \& Violence- Provoking Speech Classifiers}
\label{sec:hate_and_danger_classifiers}
We now focus on developing state-of-the-art classifiers to assess their ability to detect violence-provoking and hateful speech targeting Asian communities (\textit{RQ2}). We evaluate and contrast both fine-tuned BERT-based language models as well as large language models like Mixtral~\cite{jiang2024mixtral} for the following two tasks: \textit{(i)} hateful and \textit{(ii)} violence-provoking speech detection.

\vspace{0.01in}
\noindent\textbf{Hateful speech detection}:
% \gaurav{continue from here, cite PLMs}
We consider the $954$ Twitter posts identified as hateful in our community-annotated dataset as positive examples for this task. For negative examples, we augment the $44$ `not hateful' examples in our dataset by adding the `neutral' and `counter-hate' Twitter posts curated by ~\citeauthor{he2021racism} (\citeyear{he2021racism}), resulting in a total of $1861$ `not hateful' examples. We then split the dataset into an $80:20$ ratio to create the training and validation set. In Table \ref{tab:performance-metrics-hateful}, we report the macro-averaged classification metrics, averaged over $5$ different runs, for models that span fine-tuned language models (DistilBERT~\cite{sanh2019distilbert}, BERT~\cite{devlin2018bert}, and RoBERTa~\cite{liu2019roberta}), and zero-shot and few-shot LLMs -- Flan T5-XL~\cite{chung2022scaling} and Mixtral-Ins~\cite{jiang2024mixtral}. 

\noindent\textbf{Implementation details}: The process of fine-tuning BERT-based language models involves replacing the ``pre-training head'' of the model with a randomly initialized ``classification head''. The randomly initialized parameters in the classification head are learned by \textit{fine-tuning} the model on classification examples while minimizing the cross-entropy loss. To train the models, we use Adam optimizer~\cite{kingma2014adam} with a learning rate initialized at $10^{-4}$, with a batch size of $16$ and default hyper-parameters~\cite{wolf2020transformers}. We used early stopping to stop training when the loss value on the validation set stops to improve for 3 consecutive epochs. For large language models, we carefully crafted the task-specific instructions (shown in Appendix \ref{sec:mixtral-ins}) to obtain zero-shot outputs and provided 16 randomly-sampled examples (8 per class) along with the instruction to experiment with the few-shot learning setting.

Table \ref{tab:performance-metrics-hateful} shows that all the classifiers are effective in identifying hateful speech, consistently demonstrating a $F_1$ score greater than $0.80$. Few-shot learning with Mixtral-Ins demonstrates the best performance among all the models, with an $F_1$ score of $0.89$. The performance of the RoBERTa-large model, albeit fine-tuned using all the training set, is also competitive. In effect, like prior studies~\cite{he2021racism}, we find that state-of-the-art NLP classifiers can effectively discern anti-Asian hateful speech from not hateful speech.

\vspace{0.01in}
\noindent\textbf{Violence-provoking speech detection:} We consider $246$ Twitter posts identified as violence-provoking in our community-annotated dataset as positive examples from this task. For negative examples, we combine the hateful but not violence-provoking examples and the not hateful from our dataset, the neutral and counter-hate examples from ~\citeauthor{he2021racism} (\citeyear{he2021racism}), resulting in a total of $2034$ negative examples. We again split the dataset into an $80:20$ ratio to create the training and validation set. We show the classification results in Table \ref{tab:performance-metrics-dangerous}. We adopt the same models and training strategies as those for developing hateful classifiers.

Table \ref{tab:performance-metrics-dangerous} shows that RoBERTa-large is most effective in detecting violence-provoking speech and achieves an $F_1$ score of $0.69$. The performance of the classifier is lacking, especially in comparison to the hateful classifier, indicating the difficulty in detecting violence-provoking speech. Additionally, few-shot learning using Mixtral-Ins demonstrates a notably subpar performance in comparison to fine-tuned RoBERTa, with an $F_1$ score of $0.62$, indicating that a limited examples may not be enough to model the variability and subjectivity of violence-provoking speech. Next, we perform an error analysis to understand the underlying challenges. 

\begin{table}[!t]
\centering
\resizebox{0.50\textwidth}{!}{%
\begin{tabular}{l c c c c c}
\toprule
\textbf{Model} & \textbf{\# Parameters} & \textbf{F1} & \textbf{Precision} & \textbf{Recall} & \textbf{Accuracy} \\ \midrule
Random (uniform) & -- & 0.5071 & 0.5151 & 0.5162 & 0.5147 \\ \midrule
Random (biased) & -- & 0.5021 & 0.5023 & 0.5023 & 0.5295 \\ \midrule
DistilBERT-base-uncased & 66M & 0.8559 & 0.8489 & 0.8520 & 0.8620 \\ \midrule
BERT-large-uncased & 340M & 0.8833 & 0.8737 & 0.8818 & 0.8773 \\ \midrule
RoBERTa-large & 354M & 0.8897 & 0.8762 & 0.8871 &  0.8807 \\ 
\midrule
% GPT-2 & 1.5B & 0.8354 & 0.8445 & 0.8291 & 0.8491 \\ 
% \midrule
Flan T5-xl (Ins + n = 16) & 3B & 0.8192 & 0.8153 & 0.8254 & 0.8278 \\
\midrule
Mixtral-Ins (zero-shot) & 8x7B & 0.8642 & 0.8596 & 0.8613 & 0.8742 \\ \midrule
Mixtral-Ins (Ins + n = 16) & 8x7B & 0.8941 & 0.8838 & 0.8916 & 0.8893 \\

\bottomrule
\end{tabular}%
}
\caption{{Classification performance of models trained to detect \textbf{hateful} speech. The values are averages of $5$ experimental runs with different random seeds.}}
\label{tab:performance-metrics-hateful}
\end{table}

\vspace{0.01in}
\noindent\textbf{Error Analysis of the Violence-Provoking Speech Classifier:}
We aim to discern the limited capabilities of the above classifiers to detect violence-provoking content. First, we note that the negative examples include hateful but not violence-provoking content, and counter- and neutral speech. While the model can distinguish violence-provoking content from counter- and neutral speech, it struggles to distinguish hateful but not violence-provoking content from violence-provoking content. This is primarily because violence-provoking speech is a subset of hateful speech that involves more nuanced expressions.\footnote{{An alternative formulation for detecting violence-provoking speech could have been a 3-way categorization of content into `violence-provoking', `hateful', and `other' categories. However, we found that this formulation also leads to limited distinguishability between hateful and violence-provoking categories (macro $F_1$ score of $0.63$ with a RoBERTa-large classifier), with the majority of miscategorizations being among the two harmful categories.}} 

\vspace{0.01in}
\noindent\textit{{Violence-provoking and Not-violence-provoking examples demonstrate statistical similarities}}: Our experiments indicate that hateful but not violence-provoking ($n = 173$) and violence-provoking ($n = 246$) speech demonstrate statistically indistinguishable sentiment scores (quantified using VADER~\cite{hutto2014vader}), positive or negative emotions (quantified using `posemo' and `negemo' categories in LIWC~\cite{tausczik2010psychological}), and swear words (quantified using `swear' category). The p-values computed using a two-sample t-test with equal variances assumption were $> 0.05$ in all the cases. 

\vspace{0.01in}
\noindent\textit{{Lack of BERT-based models to effectively encode compositionality}}: Due to the lack of statistical differences in occurrence-based identifiers, effective modeling of the \textit{compositionality} in language becomes crucial. However, it has been demonstrated that pre-trained language models like RoBERTa and BERT struggle with compositional semantics (for instance, negation and Semantic Role Labeling)~\cite{staliunaite2020compositional}. We illustrate this using concrete examples in Appendix \ref{sec:examples_compositionality}. 

Our work highlights the challenges in detecting violence-provoking speech and the need for stronger language modeling capabilities beyond fine-tuned pre-trained language models or employing LLMs that learn with few examples.

\begin{table}[!t]
\centering
\resizebox{0.50\textwidth}{!}{%
\begin{tabular}{l c c c c c}
\toprule
\textbf{Model} & \textbf{\# Parameters} & \textbf{F1} & \textbf{Precision} & \textbf{Recall} & \textbf{Accuracy} \\ \midrule
Random (uniform) & -- & 0.4302 & 0.5025 & 0.5059 & 0.5307 \\ \midrule
Random (biased) & -- & 0.4745 & 0.4723 & 0.4775 & 0.8032 \\ \midrule
DistilBERT-base-uncased & 66M & 0.6772 & 0.6843 & 0.6790 & 0.8576 \\ \midrule
BERT-large-uncased & 340M & 0.6845 & 0.6912 & 0.6897 & 0.8603 \\ \midrule
RoBERTa-large & 354M & 0.6975 & 0.7007 & 0.6991 & 0.8689 \\ 
\midrule
Flan T5-xl (Ins + n = 16) & 3B & 0.5214 & 0.5317 & 0.5249 & 0.6556 \\
\midrule
Mixtral-Ins (zero-shot) & 8x7B & 0.5722 & 0.5612 & 0.5652 & 0.7104 \\ \midrule
Mixtral-Ins (Ins + n = 16) & 8x7B & 0.6213 & 0.5965 & 0.6241 & 0.7743 \\

\bottomrule
\end{tabular}%
}
\caption{{Classification performance of models trained to detect \textbf{violence-provoking} speech. The values are averages of $5$ runs with different random seeds.}}
\label{tab:performance-metrics-dangerous}
\end{table}

\section{Discussion and Conclusion}
\label{sec:discussion_conclusion}
We developed a comprehensive codebook to enable the conceptual identification of violence-provoking speech and distinguish it from more subjective hateful speech (\textit{RQ1}). We then used the codebook to obtain annotations from Asian community members. The high inter-annotator agreement scores demonstrate the effectiveness of our codebook and the quality of the collected data. We then used the annotated data to train classifiers that can be used for detecting hateful and violence-provoking speech. We highlighted the lacking capabilities of NLP classifiers in effectively distinguishing violence-provoking speech from hateful speech and conducted error analysis to aid future research (\textit{RQ2}). 
% We also find that dangerous posts are more likely to contain misinformation than hateful posts, hinting at the greater influence of misinformative narratives in a more severe form of harmful speech (\textit{RQ3}). The higher prevalence of misinformation in dangerous speech than in hateful speech is consistent across large-scale machine-labeled datasets as well as human-annotated datasets.

\vspace{0.01in}
\noindent\textbf{Implications}: %\gaurav{updated} 
We believe that the findings from our study can enable informed decision-making by different stakeholders: \textit{(a)} policy-makers who are responsible for regulating harmful speech, including both hateful speech and violence-provoking speech, \textit{(b)} practitioners who are responsible for algorithmic accountability of online platforms, and \textit{(c)} the targeted Asian community members. Our work hints at the need for a tiered penalty system on online platforms that may allow for more nuanced and proportionate responses to varying types of harmful content, enhancing algorithmic accountability. Tiered penalties are also more fair as they align penalties with the severity of the offense, thereby offering a balanced deterrent that could encourage more thoughtful online interaction. Furthermore, our study underscores the importance of tailored, trauma-informed interventions~\cite{han2021trauma} to support targeted communities as there is a need to create more holistic and humane approach to protecting targeted communities.

\section{Limitations \& broader perspective} 

\noindent\textit{Limitations and future work}: It is important to be clear about the limitations of this study. Our study focuses on Twitter posts explicitly mentioning anti-Asian keywords in the context of COVID-19. To avoid moderation penalties, users often refer to the targeted communities as `they' or `them,' which would have been skipped in our study. However, considering such posts requires additional data curation efforts and may lead to more noise in the samples and a lower density of violence-provoking posts. 
In future work, we will conduct a user-profile level analysis to uncover the individual-level traits and content exposure that may trigger violence-provoking expressions. We also intend to leverage the developed codebook and community-crowdsourced dataset to develop more effective approaches to detect violence-provoking speech.

\vspace{0.05in}
\noindent\textit{Dataset and resources}: The data collection and annotation for this study have been approved by the Institutional Review Board (IRB) at the researchers' institution. The annotators were informed about the potentially hateful and violence-provoking nature of the content, targeting individuals from the same ethnicity (i.e., Asians), and had the agency to discontinue participation at any point. We anonymized all data, replaced all user mentions with `@user', and rephrased all examples in the paper to avoid traceability. The data was stored on IRB-approved devices. The resources developed in this study are available at \url{https://claws-lab.github.io/violence-provoking-speech/}. The existing resources (models, data, and software) we used are publicly available for research, and we abide by their terms of use. 

\vspace{0.05in}
\noindent\textit{Broader social impact}: The limitation of machine learning models in accurately distinguishing between violence-provoking and hateful speech could lead to false positives, unfairly penalizing benign users, if not used responsibly. Moreover, we condemn any reinforcement of harmful stereotypes or prejudices against the Asian community that could be inadvertently caused by the presented examples of Twitter posts. Some of the authors of this work identify as Asians, which enabled us to contextualize our findings and discussions with their lived experiences. Some of the content included in the paper could be offensive, especially to readers of Asian descent, for which the authors suggest caution to the readers.

\section{Acknowledgements}

This research/material is based upon work supported in part by
NSF grants CNS-2154118, ITE-2137724, ITE-2230692, CNS2239879, Defense Advanced Research Projects Agency (DARPA) under Agreement No. HR00112290102 (subcontract No. PO70745), CDC, and funding from Microsoft and the American Foundation for Suicide Prevention (AFSP). Any opinions, findings, and conclusions or recommendations expressed in this material are those of the author(s) and do not necessarily reflect the position or policy of DARPA, DoD, SRI International, CDC, NSF, NIH, AFSP, and no official endorsement should be inferred. Gaurav is partially supported by the JP Morgan AI Research PhD Fellowship and the Snap Research Fellowship.  We are thankful to Kefai (KeKe) Debebe (CDC) and Daniel A. Bowen (CDC) for their continuous feedback on this work. We are also grateful to Dr. Morgan Clark (ADL) for their support and feedback while developing the codebook.

\bibliography{anthology,custom}
\bibliographystyle{acl_natbib}

\appendix

\section{Appendix}

\subsection{Instruction for Mixtral-Ins and Flan-T5-XL}
\label{sec:mixtral-ins}
We use the format depicted in the subsequent paragraphs for specifying the instructions and providing few-shot examples to the large language models.

\vspace{0.03in}
\noindent\textbf{Hateful speech detection}:
We use the definition used by United Nations to guide the LLMs to detect hatespeech. The definition is available here: \url{https://www.un.org/en/hate-speech/understanding-hate-speech/what-is-hate-speech}.

\vspace{0.03in}
\noindent\textbf{Violence-provoking speech detection}:
We use the definition based on our developed codebook, which is inspired by the work of ~\citet{benesch2021dangerous} and operationalized in the context of anti-Asian Twitter posts during COVID-19; see Table \ref{tab:codebook} for the detailed codebook.

\subsection{Role of modeling compositionality in language for violence-provoking speech detection}
\label{sec:examples_compositionality}
We demonstrate using qualitative examples how BERT-based models struggle to detect violence-provoking speech, possibly, due to their lacking abilities to model compositionality in language. Consider the following examples: ``\textit{these f**king ch**ks brutally eat dogs and all nasty animals that's why squint all the time}'' is labeled as violence-provoking because it portrays the community as cruel, repulsive, and uncivilized but is incorrectly predicted to be `not violence-provoking'. However, ``\textit{i hate f**king ch**ks they are just disgusting and don't see without squinting}'' is labeled as hateful but incorrectly predicted to be `violence-provoking'. Similarly, ``\textit{they steal our jobs, make the economy suffer, spread diseases, and we welcome them with open hands. these ch**ks need to pay}'' (violence-provoking) and ``\textit{the virus has triggered an economic downturn, millions killed and so many jobs lost. we should have closed our borders to stop the ch**ks in time}'' (not violence-provoking) are similar at a superficial level but convey different meanings that materialize due to the compositionality in language -- however, they both are predicted to be `violence-provoking' by the classifier.

\begin{tcolorbox}[colback=gray!10, % Light gray background
                  colframe=gray, % Gray frame color
                  arc=1mm, % Rounded corners
                  boxsep=0.1mm, % Padding between text and box
                  boxrule=0.2pt] % Frame thickness
{You are an AI assistant that helps identify hateful speech targeting Asian community members.\\~\\
Use the following guidelines to identify whether a given Twitter post is hateful.\\~\\
Hate speech is defined as attacks or use of pejorative or discriminatory language with reference to a person or a group on the basis of who they are, in other words, based on their religion, ethnicity, nationality, race, colour, descent, gender or other identity factor. Consider the following examples:\\~\\

Tweet: \texttt{<example 1>}\\
Label: \texttt{<label 1>}\\
...\\
Tweet: \texttt{<example N>}\\
Label: \texttt{<label N>}\\~\\
Only respond using the following strings: `not hateful' or `hateful'. Do NOT respond with anything else.\\~\\

Tweet: \texttt{<current example to label>}\\
Label:
}
\end{tcolorbox}

\begin{tcolorbox}[colback=gray!10, % Light gray background
                  colframe=gray, % Gray frame color
                  arc=1mm, % Rounded corners
                  boxsep=0.1mm, % Padding between text and box
                  boxrule=0.2pt] % Frame thickness
{ You are an AI assistant that helps identify violence-provoking speech targeting Asian community members.\\~\\
Use the following guidelines to identify whether a given Twitter post is violence-provoking.\\~\\
Violence-provoking speech is defined as community-targeting speech that could lead to violence against the community member(s). It could either be explicit or implicit. Implicit forms of such speech indirectly provokes violence using elements like dehumanization, guilt attribution, threat construction, virtuetalk, and constructing future-bias against the community member(s). Violence-provoking speech targets the member(s) of the general public and not political entities, countries, or inanimate objects. Consider the following examples:\\~\\

Tweet: \texttt{<example 1>}\\
Label: \texttt{<label 1>}\\
...\\
Tweet: \texttt{<example N>}\\
Label: \texttt{<label N>}\\~\\
Only respond using the following strings: `not violence-provoking' or `violence-provoking'. Do NOT respond with anything else.\\~\\

Tweet: \texttt{<current example to label>}\\
Label:
}
\end{tcolorbox}

\begin{table*}[!t]
    \centering
    \definecolor{headercolor}{rgb}{0.9, 0.9, 0.9}
    \resizebox*{!}{0.55\textwidth}{
    \small
    \begin{tabular}{ p{0.24\linewidth} p{0.16\linewidth} p{0.17\linewidth} p{0.71\linewidth} }
        \toprule
         \rowcolor{headercolor} \textbf{Concept} & \textbf{Sub-concept} & \textbf{Identifiers} & \textbf{Definition} \\ \toprule
        \multirow{7}{*}{\parbox{0.22\textwidth}{Community-targeting violence-provoking speech could lead to violence against the community}} & Direct violence-provoking & Explicit mention of violence & Directly mention attacking the members of the targeted community and causing them physical harm \\ \cmidrule{2-4}
         & \multirow{2}{*}{\parbox{0.13\textwidth}{Indirect violence-provoking}} & Dehumanization & Dehumanization is the description of other people (in this case, Asians) as something other than human or less than human. This can involve likening them to bacteria, insects, or other repulsive or unwanted creatures \\ \cmidrule{3-4}
         & & Guilt Attribution & Victims are often deemed guilty as a group, deserving collective punishment for the specific crimes of some of their “members.”\\ \cmidrule{3-4}
         & & Threat construction & ... asserts that the in-group faces serious and often mortal threats from the victims to-be, which makes violence seem defensive, and therefore proper and necessary \\ \cmidrule{3-4}
         & & Prediction of Violence & Violence is presented as inevitable and necessary as a way to protect the in-group from harm or annihilation  \\ \cmidrule{3-4}
         & & Virtuetalk & The valorization of violence by associating it with a range of praiseworthy characteristics, and the parallel denigration of resistance or non-participation as indicating a lack of proper character traits, a deplorable “weakness,” or a range of other deficiencies.\\ \cmidrule{3-4}
         & & Future-bias & The confident anticipation of future goods that will be accrued through violence, and which are so extensive and so enduring in a relatively certain future that they easily outweigh the moral costs of victims’ deaths in the here and now.  \\ \midrule
         Targets member(s) of the general public and not political entities, organizations, countries, or inanimate objects & & & The speech targets one or many members of the public that belong to Asian communities and not political entities, political personalities, countries, organizations, or inanimate objects like (applications, products, food items, etc.). \\ \midrule
         Speech is aimed to harm and does not (i) counter violence-provoking speech, (ii) share lived or witnessed experiences, and (iii) comprise news articles or reports of attacks
          & & & The speech is intended to direct harm toward the Asian communities and does not mention narratives that include violence-provoking speech as a means to bring attention to them. This would exclude sharing lived or witnessed experiences and new articles that report incidents \\ \bottomrule
    \end{tabular}}
    \caption{{\textbf{Codebook for anti-Asian violence-provoking speech with definitions of underlying concepts and identifiers.} We consider aspects related to both implicit and explicit forms of violence-provoking speech against Asian community members.}  
    }
    \label{tab:codebook}
\end{table*}

\begin{table*}[!b]
    \centering
    \resizebox{0.8\textwidth}{!}{%
    \begin{tabular}{p{0.19\linewidth} p{0.79\linewidth}}
    \toprule
    \rowcolor{headercolor} \textbf{Set} & \textbf{Keywords}\\
    \toprule
    COVID-19 &  `coronavirus', `covid 19', `covid-19', `covid19', `corona virus', `virus'\\ 
    \toprule
    Anti-Asian & `chink', `chinky', `chonky', `churka', `cina', `cokin', `coolie', `dink', `niakoue', `pastel de flango', `slant', `slant eye', `slopehead', \textcolor{bondiblue}{`slope head'}, `ting tong', `yokel', \textcolor{bondiblue}{`pasteldeflango'},
                                \textcolor{bondiblue}{`slanteye'}, \textcolor{bondiblue}{`slitty'}, \textcolor{bondiblue}{`squinty'}, \textcolor{bondiblue}{`kungflu'}, \textcolor{bondiblue}{`gooks'}, \textcolor{bondiblue}{`churka'}, \textcolor{bondiblue}{`wuflu'}, \textcolor{bondiblue}{`antichinazi'}, \textcolor{bondiblue}{`slanty'}, \textcolor{bondiblue}{`kungfuflu'}, \textcolor{bondiblue}{`squint'}, \textcolor{bondiblue}{`gook'}, \textcolor{bondiblue}{`slanted'}, \textcolor{bondiblue}{`niakouee'}, \textcolor{bondiblue}{`chinks'}\\
    \bottomrule
    \end{tabular}%
    }
    \caption{Final set of keywords used for Twitter data collection. The original keywords are shown in black, and the expanded ones are in \textcolor{bondiblue}{blue}.}
    \label{tab:final_keywords}
\end{table*}

\begin{table*}[!b]
    \centering
    \definecolor{headercolor}{rgb}{0.9, 0.9, 0.9}
    \resizebox{0.92\textwidth}{!}{%
    \begin{tabular}{p{0.14\linewidth} p{1.1\linewidth}}
    \toprule
    \rowcolor{headercolor} \textbf{Set} & \textbf{Keywords}\\
    \toprule
    \textbf{Initial} &  `to eat', `to wash', `cockroaches', `microbes', `parasites', `yellow ants', `logs', `enemy morale', `devils', `satan', `demons', `weak', `fanatic', `mercy', `savage'
\\ 
    \toprule
    \textbf{Expanded} & `cut', `sterilize', `exorcise', `purify', `rein', `feeble', `peel', `undermine', `chew', `bigot', `weaken', `termites', `soak', `germs', `use', `rough', `viruses', `resist', `satan', `cockroaches', `launder', `mercy', `eradicate', `defeat', `demons', `fanatic', `brutal', `parasites', `ticks', `exterminate', `squash', `bugs', `stumps', `the prince of darkness', `vicious', `vanquish', `evil spirits', `fire ants', `chop', `yellow ants', `hellions', `tapeworms', `ogres', `fortify', `frail', `the serpent', `intolerant', `pity', `puritan', `roaches', `spunk', `savor', `delicate', `ferocious', `esprit de corps', `monomaniac', `devils', `guts', `pests', `clemency', `destabilize', `kill', `lice', `helminths', `vulnerable', `ants', `feast', `savage', `expel', `primitive', `gobble', `fleas', `beetles', `to eat', `consume', `vermin', `nibble', `beasts', `charity', `mycoplasmas', `rid', `dissolve', `leeches', `creatures', `zealot', `antennae', `erode', `enemy morale', `nematodes', `microbes', `destroy', `fierce', `defenseless', `uncivilized', `fundamentalist', `extremist', `gorge', `diminish', `leniency', `powerless', `purge', `yeast', `untamed', `to wash', `monsters', `eliminate', `rinse', `spiders', `mites', `fiends', `incapacitated', `infirm', `dispose', `protozoa', `devour', `bark', `fungi', `wild', `beef', `stiffen', `weak', `impotent', `forgiveness', `banish', `fragile', `insects', `virions', `timbers', `lucifer', `barbaric', `outwit', `lower', `pathogens', `remove', `reduce', `bacteria', `grace', `devil', `evil', `adversary`, `demon'\\
    \bottomrule
    \end{tabular}%
    }
    \caption{Initial and expanded keywords used to find Twitter posts with a high concentration of violence-provoking expressions.}
    \label{tab:dangerous_keywords}
\end{table*}

\balance

\end{document}